\documentclass[11pt,a4paper]{article}
\usepackage[hyperref]{eacl2021}
\usepackage{times}
\usepackage{latexsym}

\usepackage{microtype}

\aclfinalcopy 
\def\aclpaperid{388} 

\usepackage{amsmath}
\usepackage{graphicx}
\usepackage{subcaption}
\usepackage{booktabs}
\usepackage{multirow}
\usepackage{makecell}
\usepackage{array}
\newcolumntype{x}[1]{>{\centering\arraybackslash\hspace{0pt}}p{#1}}
\newcolumntype{y}[1]{>{\arraybackslash\hspace{0pt}}p{#1}}

\newcommand{\mbf}[1]{\mathbf{#1}}

\title{Generating Syntactically Controlled Paraphrases \\ \vspace{0.1em} without Using Annotated Parallel Pairs}

\author{Kuan-Hao Huang \\
  University of California, Los Angeles \\
  \texttt{khhuang@cs.ucla.edu} \\\And
  Kai-Wei Chang \\
  University of California, Los Angeles \\
  \texttt{kwchang@cs.ucla.edu} \\}

\date{}

\begin{document}
\maketitle

\begin{abstract}
Paraphrase generation plays an essential role in natural language process (NLP), and it has many downstream applications. However, training supervised paraphrase models requires many annotated paraphrase pairs, which are usually costly to obtain. On the other hand, the paraphrases generated by existing unsupervised approaches are usually syntactically similar to the source sentences and are limited in diversity. In this paper, we demonstrate that it is possible to generate syntactically various paraphrases without the need for annotated paraphrase pairs. We propose \emph{Syntactically controlled Paraphrase Generator} (SynPG), an encoder-decoder based model that learns to disentangle the semantics and the syntax of a sentence from a collection of unannotated texts. The disentanglement enables SynPG to control the syntax of output paraphrases by manipulating the embedding in the syntactic space. Extensive experiments using automatic metrics and human evaluation show that SynPG performs better syntactic control than unsupervised baselines, while the quality of the generated paraphrases is competitive. We also demonstrate that the performance of SynPG is competitive or even better than supervised models when the unannotated data is large. Finally, we show that the syntactically controlled paraphrases generated by SynPG can be utilized for data augmentation to improve the robustness of NLP models.
\end{abstract}

\section{Introduction}
Paraphrase generation \cite{McKeown83rules} is a long-lasting task in natural language processing (NLP) and has been greatly improved by recently developed machine learning approaches and large data collections. Paraphrase generation demonstrates the potential of machines in semantic abstraction and sentence reorganization and has already been applied to many NLP downstream applications, such as question answering \cite{Yu18appqa}, chatbot engines \cite{Yan16appcb}, and sentence simplification \cite{Zhao18appss}.

In recent years, various approaches have been proposed to train sequence-to-sequence (seq2seq) models on a large number of annotated paraphrase pairs \cite{Prakash16seq2seq1,Lapata17seq2seq2,Cao17seq2seq3, Egonmwan19seq2seq4}. Some of them control the syntax of output sentences to improve the diversity of paraphrase generation \cite{Iyyer18scpn,Goyal20preorder,Kumar20sgcp}. However, collecting annotated pairs is expensive and induces challenges for some languages and domains. On the contrary, unsupervised approaches build paraphrase models without using parallel corpora \cite{Li18rl,Roy19vqvae,Zhang19sivae}. Most of them are based on the variational autoencoder \cite{Bowman16vae} or back-translation \cite{Lapata17seq2seq2,Gimpel18paranmt,Hu19parabank}. Nevertheless, without the consideration of controlling syntax, their generated paraphrases are often similar to the source sentences and are not diverse in syntax. 

\begin{figure}[t]
	\centering
    \includegraphics[width=.99\columnwidth]{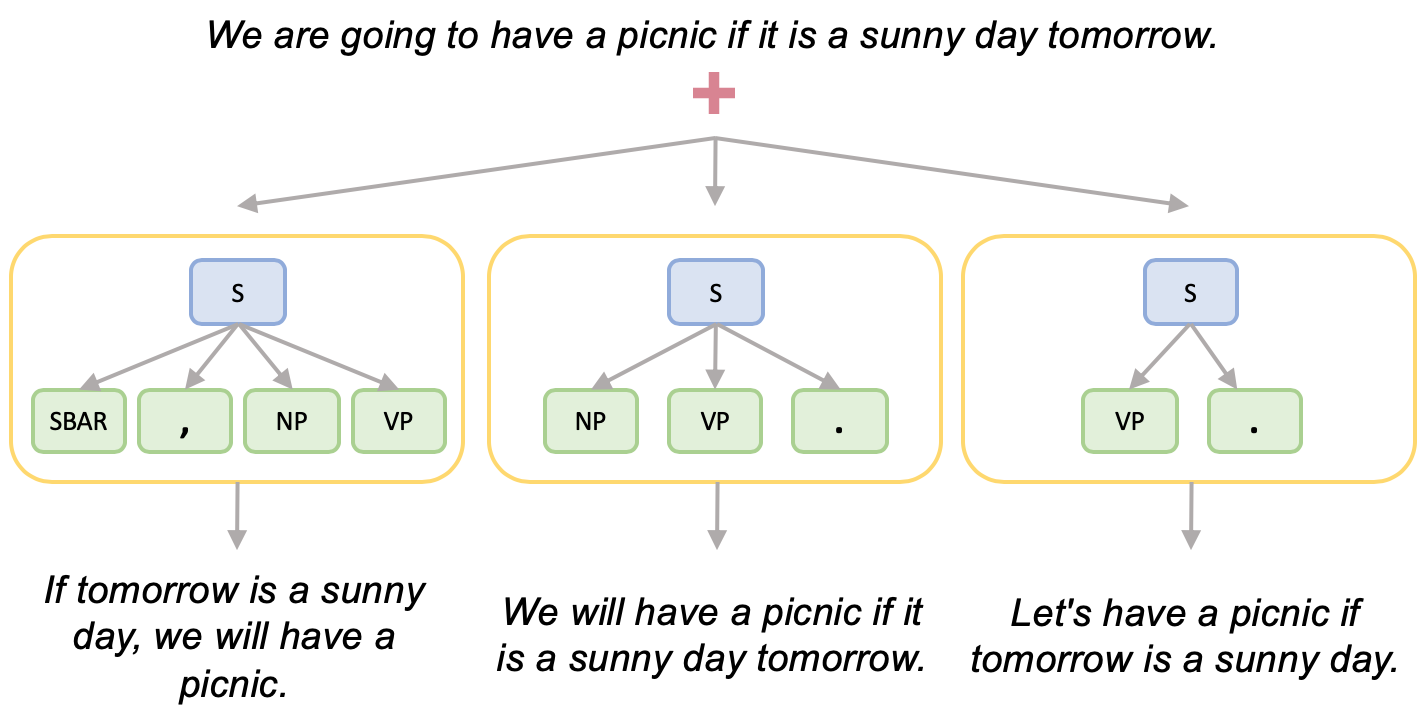}
    \caption{Paraphrase generation with syntactic control. Given a source sentence and a target syntactic specification (either a full parse tree or top levels of a parse tree), the model is expected to generate a paraphrase with the syntax following the given specification.}
    \label{fig:intro}
    \vspace{-1.2em}
\end{figure}

This paper presents a pioneering study on syntactically controlled paraphrase generation based on disentangling semantics and syntax. We aim to disentangle one sentence into two parts: 1) the semantic part and 2) the syntactic part. The semantic aspect focuses on the meaning of the sentence, while the syntactic part represents the grammatical structure. When two sentences are paraphrased, their semantic aspects are supposed to be similar, while their syntactic parts should be different. To generate a syntactically different paraphrase of one sentence, we can keep its semantic part unchanged and modify its syntactic part. 

Based on this idea, we propose \textbf{Syn}tactically Controlled \textbf{P}araphrase \textbf{G}enerator (SynPG)\footnote{Our code and the pretrained models are available at \url{https://github.com/uclanlp/synpg}}, a Transformer-based model \cite{Vaswani17transformer} that can generate syntactically different paraphrases of one source sentence based on some target syntactic parses. SynPG consists of a semantic encoder, a syntactic encoder, and a decoder. The semantic encoder considers the source sentence as a bag of words without ordering and learns a contextualized embedding containing only the semantic information. The syntactic encoder embeds the target parse into a contextualized embedding including only the syntactic information. Then, the decoder combines the two representations and generates a paraphrase sentence. The design of disentangling semantics and syntax enables SynPG to learn the association between words and parses and be trained by reconstructing the source sentence given its unordered words and its parse. Therefore, we do not require any annotated paraphrase pairs but only unannotated texts to train SynPG.

We verify SynPG on four paraphrase datasets: ParaNMT-50M \cite{Gimpel18paranmt}, Quora \cite{Iyer17quora}, PAN \cite{Madnani12pan}, and MRPC \cite{Dolan04mrpc}. The experimental results reveal that when being provided with the syntactic structures of the target sentences, SynPG can generate paraphrases with the syntax more similar to the ground truth than the unsupervised baselines. The human evaluation results indicate that SynPG achieves competitive paraphrase quality to other baselines while its generated paraphrases are more accurate in following the syntactic specifications. In addition, we show that when the training data is large enough, the performance of SynPG is competitive or even better than supervised approaches. Finally, we demonstrate that the syntactically controlled paraphrases generated by SynPG can be used for data augmentation to defense syntactically adversarial attack \cite{Iyyer18scpn} and improve the robustness of NLP models.

\section{Unsupervised Paraphrase Generation}
\label{sec:method}

We aim to train a paraphrase model without using annotated paraphrase pairs. Given a source sentence $\mbf x = (x_1, x_2, ..., x_n) $, our goal is to generate a paraphrase sentence $\mbf y = (y_1, y_2, ..., y_m)$ that is expected to maintain the same meaning of $\mbf x$ but has a different syntactic structure from $\mbf x$. 

\paragraph{Syntactic control.}
Motivated by previous work \cite{Iyyer18scpn,Zhang19sivae,Kumar20sgcp}, we allow our model to access additional syntactic specifications as the control signals to guide the paraphrase generation. More specifically, in addition to the source sentence $\mbf x$, we give the model a target constituency parse $\mbf p$ as another input. Given the input $(\mbf x, \mbf p)$, the model is expected to generate a paraphrase $\mbf y$ that is semantically similar to the source sentence $\mbf x$ and syntactically follows the target parse $\mbf p$. In the following discussions, we assume the target parse $\mbf p$ to be a full constituency parse tree. Later on, in Section~\ref{sec:pg}, we will relax the syntax guidance to be a \emph{template}, which is defined as the top two levels of a full parse tree. We expect that a successful model can control the syntax of output sentences and generate syntactically different paraphrases based on different target parses, as illustrated in Figure~\ref{fig:intro}.

Similar to previous work \cite{Iyyer18scpn,Zhang19sivae}, we linearize the constituency parse tree to a sequence. For example, the linearized parse of the sentence ``\textit{He eats apples.}'' is \texttt{(S(NP(PRP))(VP(VBZ)(NP(NNS)))(.))}. Accordingly, a parse tree can be considered as a sentence $\mbf p = (p_1, p_2, ..., p_k)$, where the tokens in~$\mbf p$ are non-terminal symbols and parentheses.

\begin{figure*}[!t]
	\centering
    \includegraphics[width=.75\textwidth]{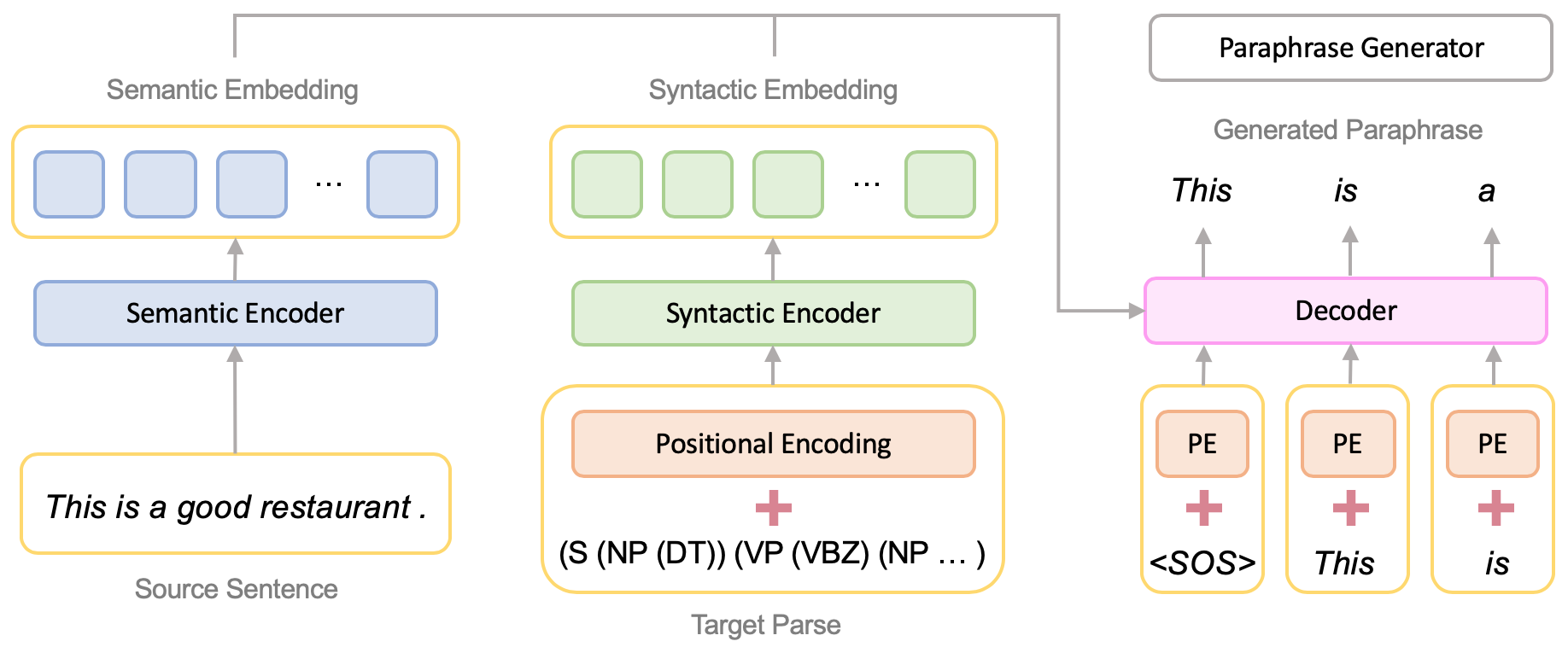}
    \caption{SynPG embeds the source sentence and the target parse into a semantic embedding and a syntactic embedding, respectively. Then, SynPG generates a paraphrase sentence based on the two embeddings.}
    \label{fig:pg}
    \vspace{-1em}
\end{figure*}

\subsection{Proposed Model}
Our main idea is to disentangle a sentence into the semantic part and the syntactic part. Once the model learns the disentanglement, it can generate a syntactically different paraphrase of one given sentence by keeping its semantic part unchanged and modifying only the syntactic part.

Figure~\ref{fig:pg} illustrates the proposed paraphrase model called SynPG, a seq2seq model consisting of a semantic encoder, a syntactic encoder, and a decoder. The semantic encoder captures only the semantic information of the source sentence~$\mbf x$, while the syntactic encoder extracts only the syntactic information from the target parse $\mbf p$. The decoder then combines the encoded semantic and syntactic information and generates a paraphrase $\mbf y$. We discuss the details of SynPG in the following.

\paragraph{Semantic encoder.} The semantic encoder embeds a source sentence $\mbf x$ into a contextualized semantic embedding $\mbf z_{sem}$. In other words,
\vspace{-0.5em}
\[ \mbf z_{sem} = (z_1, z_2, ..., z_n) = \text{Enc}_{sem}((x_1, x_2, .., x_n)).
\vspace{-0.5em}\]

The semantic embedding $\mbf z_{sem}$ is supposed to contain only the semantic information of the source sentence $\mbf x$. To separate the semantic information from the syntactic information, we use a Transformer \cite{Vaswani17transformer} \emph{without the positional encoding} as the semantic encoder. We posit that by removing position information from the source sentence $\mbf x$, the semantic embedding $\mbf z_{sem}$ would encode less syntactic information.

We assume that words without ordering capture most of the semantics of one sentence. Indeed, semantics is also related to the order. For example, exchanging the subject and the object of a sentence changes its meaning. However, the decoder trained on a large corpus also captures the \emph{selectional preferences} \cite{Katz1963selectional, Wilks75selectional} in generation, which enables the decoder to infer the proper order of words. In addition, we observe that when two sentences are paraphrased, they usually share similar words, especially those words related to the semantics. For example, ``\emph{What is the best way to improve writing skills?}'' and ``\emph{How can I improve my writing skills?}'' are paraphrased, and the shared words (\emph{improve}, \emph{writing}, and \emph{skills}) are strongly related to the semantics. In Section~\ref{sec:exp}, we show that our designed semantic embedding captures enough semantic information to generate paraphrases.

\paragraph{Syntactic encoder.}
The syntactic encoder embeds the target parse $\mbf p = (p_1, p_2, ..., p_k)$ into a contextualized syntactic embedding $\mbf z_{syn}$. That is,
\vspace{-0.5em}
\[ \mbf z_{syn} = (z_1, z_2, ..., z_k) = \text{Enc}_{syn}((p_1, p_2, .., p_k)).
\vspace{-0.5em}\]

Since the target parse $\mbf p$ contains no semantic information but only syntactic information, we use a Transformer \emph{with the positional encoding} as the syntactic encoder.

\paragraph{Decoder.} 
Finally, we design a decoder that takes the semantic embedding $\mbf z_{sem}$ and the syntactic embedding $\mbf z_{syn}$ as the input and generates a paraphrase $\mbf y$. In other words,
\vspace{-0.5em}
\[ \mbf y = (y_1, y_2, ..., y_m) = \text{Dec}(\mbf z_{sem}, \mbf z_{syn}).
\vspace{-0.5em}\]

We choose Transformer as the decoder to generate $\mbf y$ autoregressively. Notice that the semantic embedding $\mbf z_{sem}$ does not encode the position information and the syntactic embedding $\mbf z_{syn}$ does not contain semantics. This forces the decoder to extract the semantics from $\mbf z_{sem}$ and retrieve the syntactic structure from $\mbf z_{syn}$. The attention weights attaching to $\mbf z_{sem}$ and $\mbf z_{syn}$ make the decoder learn the association between the semantics and the syntax as well as the relation between the word order and the parse structures. Therefore, SynPG is able to reorganize the source sentence and use the given syntactic structure to rephrase the source sentence.

\subsection{Unsupervised Training}
\label{sec:train}

Our design of the disentanglement makes it possible to train SynPG without using annotated pairs. We train SynPG with the objective to reconstruct the source sentences. More specifically, when training on a sentence $\mbf x$, we first separate $\mbf x$ into two parts: 1) an unordered word list $\bar{\mbf x}$ and 2)~its linearized parse $\mbf p_x$ (can be obtained by a pretrained parser). Then, SynPG is trained to reconstruct $\mbf x$ from $(\bar{\mbf x}, \mbf p_x)$ with the reconstruction loss
\vspace{-0.5em}
\[\mathcal{L} = - \sum\limits_{i=1}^{n} \log P(y_i=x_i | \bar{\mbf x}, \mbf p_x, \mbf y_1, ..., \mbf y_{i-1}).
\vspace{-0.5em}\]

Notice that if we do not disentangle the semantics and the syntax, and directly use a seq2seq model to reconstruct $\mbf x$ from $(\mbf x, \mbf p_x)$, it is likely that the seq2seq model only learns to copy~$\mbf x$ and ignores $\mbf p_x$ since $\mbf x$ contains all the necessary information for the reconstruction. Consequently, at inference time, no matter what target parse $\mbf p$ is given, the seq2seq model always copies the whole source sentence $\mbf x$ as the output  (more discussion in Section~\ref{sec:exp}).

On the contrary, SynPG learns the disentangled embeddings $\mbf z_{sem}$ and $\mbf z_{syn}$. This makes SynPG capture the relation between the semantics and the syntax to reconstruct the source sentence $\mbf x$. Therefore, at test time, given the source sentence $\mbf x$ and a new target parse $\mbf p$, SynPG is able to apply the learned relation to rephrase the source sentence $\mbf x$ according to the target parse $\mbf p$.

\paragraph{Word dropout.}
We observe that the ground truth paraphrase may contain some words not appearing in the source sentence; however, the paraphrases generated by the vanilla SynPG tend to include only words appearing in the source sentence due to the reconstruction training objective. To encourage SynPG to improve the diversity of the word choices in the generated paraphrases, we randomly discard some words from the source sentence during training. More precisely, each word has a probability to be dropped out in each training iteration. Accordingly, SynPG has to predict the missing words during the reconstruction, and this enables SynPG to select different words from the source sentence to generate paraphrases. More details are discussed in Section~\ref{sec:word}. 

\subsection{Templates and Parse Generator}
\label{sec:pg}

In the previous discussion, we assume that a full target constituency parse tree is provided as the input to SynPG. However, the full parse tree of the target paraphrase sentence is unlikely available at inference time. Therefore, following the setting in \citet{Iyyer18scpn}, we consider generating the paraphrase based on the \emph{template}, which is defined as the top two levels of the full constituency parse tree. For example, the template of  \texttt{(S(NP(PRP))(VP(VBZ)(NP(NNS)))(.))} is \texttt{(S(NP)(VP))(.))}. 

Motivated by \citet{Iyyer18scpn}, we train a parse generator to generate full parses from templates. The proposed parse generator has the same architecture as SynPG, but the input and the output are different. The parse generator takes two inputs: a tag sequence $\mbf {tag}_x$ and a target template $\mbf t$. The tag sequence $\mbf {tag}_x$ contains all the POS tags of the source sentence $\mbf x$. For example, the tag sequence of the sentence ``\textit{He eats apples.}'' is ``\texttt{<PRP> <VBZ> <NNS> <.>}''. Similar to the source sentence in SynPG, we do not consider the word order of the tag sequence during encoding. The expected output of the parse generator is a full parse $\tilde{\mbf p}$ whose a syntactic structure follows the target template $\mbf t$.

We train the parse generator without any additional annotations as well. Let $\mbf t_x$ be the the template of $\mbf p_x$ (the parse of $\mbf x$), we end-to-end train the parse generator with the input being $(\mbf {tag}_x, \mbf t_x)$ and the output being $\mbf p_x$.

\paragraph{Generating paraphrases from templates.}
The parse generator makes us generate paraphrases by providing target templates instead of target parses. The steps to generate a paraphrase given a source sentence $\mbf x$ and a target template $\mbf t$ are as follows:
\vspace{-0.4em}
\begin{enumerate}
\setlength{\itemsep}{-3pt}
    \item Get the tag sequence $\mbf {tag}_x$ of the source sentence $\mbf x$.
    \item Use the parse generator to generate a full parse $\tilde{\mbf p}$ with input $(\mbf {tag}_x, \mbf t)$.
    \item Use SynPG to generate a paraphrase $\mbf y$ with input $(\mbf x, \tilde{\mbf p})$.
\end{enumerate}

\paragraph{Post-processing.}
We notice that certain templates are not suitable for some source sentences and therefore the generated paraphrases are nonsensical. We follow \citet{Iyyer18scpn} and use n-gram overlap and paraphrastic similarity computed by the model\footnote{https://github.com/jwieting/para-nmt-50m} from \citet{Gimpel18paranmt} to remove nonsensical paraphrases\footnote{We set the minimum n-gram overlap to 0.3 and the minimum paraphrastic similarity to 0.7.}.

\section{Experimental Settings}
\label{sec:setup}

We conduct extensive experiments to demonstrate that SynPG performs better syntactic control than other unsupervised paraphrase models, while the quality of generated paraphrases by SynPG is comparable to others. In addition, we show that the performance of SynPG is competitive or even better than supervised models when the training data is large enough.

\subsection{Datasets}
For the training data, we consider ParaNMT-50M \cite{Gimpel18paranmt}, a paraphrase dataset containing over 50 million pairs of reference sentences and the corresponding paraphrases as well as the quality scores. We select about 21 million pairs with higher quality scores as our training examples. Notice that we use \emph{only the reference sentences} to train SynPG and unsupervised paraphrase models since we do not require paraphrase pairs.

We sample 6,400 pairs from ParaNMT-50M as the testing data. To evaluate the transferability of SynPG, we also consider the other three datasets: 1) Quora \cite{Iyer17quora} contains over 400,000 paraphrase pairs and we sample 6,400 pairs from them. 2) PAN \cite{Madnani12pan} contains 5,000 paraphrase pairs. 3) MRPC \cite{Dolan04mrpc} contains 2,753 paraphrase pairs.

\subsection{Evaluation} 
We consider paraphrase pairs to evaluate all the models. For each test paraphrase pair $(\mbf x_1, \mbf x_2)$, we consider $\mbf x_1$ as the source sentence and treat $\mbf x_2$ as the target sentence (ground truth). Let $\mbf p_2$ be the parse of $\mbf x_2$, given $(\mbf x_1, \mbf p_2)$, The model is expected to generate a paraphrase $\mbf y$ that is similar to the target sentence $\mbf x_2$.

We use BLEU score \cite{Papineni02bleu} and human evaluation to measure the similarity between $\mbf x_2$ and $\mbf y$. Moreover, to evaluate how well the generated paraphrase $\mbf y$ follows the target parse $\mbf p_2$, we define the \emph{template matching accuracy} (TMA) as follows. For each ground truth sentence~$\mbf x_2$ and the corresponding generated paraphrase $\mbf y$, we get their parses ($\mbf p_2$ and $\mbf p_y$) and templates ($\mbf t_2$ and $\mbf t_y$). Then, we calculate the percentage of pairs whose $\mbf t_y$ \emph{exactly matches} $\mbf t_2$ as the \emph{template matching accuracy.}

\begin{table*}[t]
    \centering
    \small
    \aboverulesep = 0.5mm
    \belowrulesep = 0.5mm
    \begin{tabular}{x{.14\textwidth}y{.11\textwidth}x{.06\textwidth}x{.06\textwidth}x{.06\textwidth}x{.06\textwidth}x{.06\textwidth}x{.06\textwidth}x{.06\textwidth}x{.06\textwidth}}
        \toprule
        & \multirow{2}{*}{Model} & \multicolumn{2}{c}{ParaNMT} & \multicolumn{2}{c}{Quora} & \multicolumn{2}{c}{PAN} & \multicolumn{2}{c}{MRPC} \\
        \cmidrule{3-10}
        & & TMA & BLEU & TMA & BLEU & TMA & BLEU & TMA & BLEU \\
        \midrule
        No Paraphrasing & CopyInput & 33.6 & 16.4 & 55.0 & 20.0 & 37.3 & 26.8 & 47.9 & 30.7 \\
        \midrule
        \multirow{2}{*}{\makecell{Unsupervised\\ Models}} 
        & BackTrans & 29.0 & 16.3 & 53.0 & 16.4 & 27.9 & 16.2 & 47.2 & 21.6 \\
        & VAE & 26.3 & 9.6 & 44.0 & 8.1 & 19.4 & 5.2 & 20.8 & 1.2 \\
        \midrule
        \multirow{3}{*}{\makecell{With Syntactic\\ Specifications}} & SIVAE & 30.0 & 12.8 & 48.3 & 13.1 & 26.6 & 11.8 & 21.5 & 5.1 \\
        & Seq2seq-Syn & 33.5 & 16.3 & 54.9 & 19.8 & 37.1 & \textbf{26.5} & 47.7 & \textbf{30.4} \\
        & SynPG & \textbf{71.0 }& \textbf{32.2} & \textbf{82.6} & \textbf{33.2} & \textbf{66.3} & 26.4 & \textbf{74.0} & 26.2 \\
        \bottomrule
    \end{tabular}
    \caption{Paraphrase results on four datasets. TMA denotes the template matching accuracy, which evaluates how often the generated paraphrases follow the target parses. With the syntactic control, SynPG obtains higher BLEU score and the template matching accuracy. This implies the paraphrases generated by SynPG are more similar to the ground truths and follow the target parses more accurately.}
    \label{tab:ps_sc}
\end{table*}

\begin{table*}[t]
    \centering
    \small
    \aboverulesep = 0.5mm
    \belowrulesep = 0.5mm
    \begin{tabular}{rm{.39\textwidth}m{.41\textwidth}}
        \toprule
        Model & Example 1 (ParaNMT) & Example 2 (Quora) \\
        \midrule
        Source Sent. & these children are gonna die if we don't act now. & what are the best ways to improve writing skills? \\
        Ground Truth & if we don't act quickly, the children will die. & how could i improve my writing skill? \\
        \midrule
        BackTrans & these children will die if we do not act now. & what are the best ways to improve your writing skills? \\
        VAE & these children are gonna die if we don't act now. & what are the best ways to improve writing skills? \\
        \midrule
        SIVAE & these children are gonna die if we don't act now . & what are the best ways to improve writing skills? \\
        Seq2seq-Syn & these children are gonna die if we don't act now. & what are the best ways to improve writing skills? \\
        SynPG & if we don't act now, these children will die. & how can i improve my writing skills? \\
        \bottomrule
    \end{tabular}
    \caption{Paraphrases generated by each model. SynPG can generate paraphrases with the syntax more similar to the ground truth than other baselines.}
    \vspace{-1em}
    \label{tab:ps_ex}
\end{table*}

\subsection{Models for Comparison}
We consider the following unsupervised paraphrase models: 
1)~\textbf{CopyInput}: a na\"ive baseline which directly copies the source sentence as the output without paraphrasing.
2)~\textbf{BackTrans}: back-translation is proposed to generate paraphrases  \cite{Lapata17seq2seq2,Gimpel18paranmt,Hu19parabank}. In our experiment, we use the pretrained EN-DE and DE-EN translation models\footnote{https://github.com/pytorch/fairseq/tree/master/examples/wmt19} proposed by \citet{Ng19ende} to conduct back-translation. Notice that training translation models requires additional translation pairs. Therefore, BackTrans needs more resources than ours and the translation data may not available for some low-resource languages.   
3)~\textbf{VAE}: we consider a vanilla variational autoencoder \cite{Bowman16vae} as a simple baseline.
4)~\textbf{SIVAE}: syntax-infused variational autoencoder \cite{Zhang19sivae} utilizes additional syntax information to improve the quality of sentence generation and paraphrase generation. Unlike SynPG, SIVAE does not disentangle the semantics and syntax.
5) \textbf{Seq2seq-Syn}: we train a seq2seq model with Transformer architecture to reconstruct $\mbf x$ from $(\mbf x, \mbf p_x)$ without the disentanglement. We use this model to study the influence of the disentanglement.
6) \textbf{SynPG}: our proposed model which learns disentangled embeddings.

We also compare SynPG with supervised approaches. We consider the following:
1) \textbf{Seq2seq-Sup}: a seq2seq model with Transformer architecture trained on whole ParaNMT-50M pairs.
2) \textbf{SCPN}: syntactically controlled paraphrase network \cite{Iyyer18scpn} is a supervised paraphrase model with syntactic control trained on ParaNMT-50M pairs. We use their pretrained model\footnote{https://github.com/miyyer/scpn}.

\subsection{Implementation Details}
We consider byte pair encoding \cite{Sennrich16bpe} for tokenization and use Stanford CoreNLP parser \cite{Manning14corenlp} to get constituency parses. We set the max length of sentences to 40 and set the max length of linearized parses to 160 for all the models. For the encoders and the decoder of SynPG, we use the standard Transformer \cite{Vaswani17transformer} with default parameters. The word embedding is initialized by GloVe \cite{Pennington14glove}. We use Adam optimizer with the learning rate being $10^{-4}$ and the weight decay being $10^{-5}$. We set the word dropout probability to 0.4 (more discussion in Section~\ref{sec:word}). The number of epoch for training is set to 5. 

Seq2seq-Syn, Seq2seq-Sup are trained with the similar setting. We reimplemnt VAE and SIVAE, and all the parameters are set to the default value in the original papers.

\section{Results and Discussion}
\label{sec:exp}

\subsection{Syntactic Control}
We first discuss if the syntactic specification enables SynPG to control the output syntax better. Table~\ref{tab:ps_sc} shows the template matching accuracy and BLEU score for SynPG and the unsupervised baselines. Notice that here we use the full parse trees as the syntactic specifications. We will discuss the influence of using the template as the syntactic specifications in Section~\ref{sec:tp}.

Although we train SynPG on the reference sentences of ParaNMT-50M, we observe that SynPG performs well on Quora, PAN, and MRPC as well. This validates that SynPG indeed learns the syntactic rules and can transfer the learned knowledge to other datasets. CopyInput gets high BLEU scores; however, due to the lack of paraphrasing, it obtains low template matching scores. Compared to the unsupervised baselines, SynPG achieves higher template matching accuracy and higher BLEU scores on all datasets. This verifies that the syntactic specification is indeed helpful for syntactic control. 

Next, we compare SynPG with Seq2seq-Syn and SIVAE. All models are given syntactic specifications; however, without the disentanglement, Seq2seq-Syn and SIVAE tend to copy the source sentence as the output and therefore get low template matching scores.

Table~\ref{tab:ps_ex} lists some paraphrase examples generated by all models. Again, we observe that without syntactic specifications, the paraphrases generated by unsupervised baselines are similar to the source sentences. Without the disentanglement, Seq2seq-Syn and SIVAE always copy the source sentences. SynPG is the only model can generate paraphrases syntactically similar to the ground truths.

\subsection{Human Evaluation}

We perform human evaluation using Amazon Mechanical Turk to evaluate the quality of generated paraphrases. We follow the setting of previous work \cite{Kok10hm,Iyyer18scpn,Goyal20preorder}. For each model, we randomly select 100 pairs of source sentence $\mbf x$ and the corresponding generated paraphrase $\mbf y$ from ParaNMT-50M test set (after being post-processed as mentioned in Section~\ref{sec:pg}) and have three Turkers annotate each pair. The annotations are on a three-point scale: \textbf{0} means $\mbf y$ is not a paraphrase of $\mbf x$; \textbf{1} means $\mbf x$ is paraphrased into $ \mbf y$ but $\mbf y$ contains some grammatical errors; \textbf{2} means $\mbf x$ is paraphrased into $ \mbf y$, which is grammatically correct. 

The results of human evaluation are reported in Table~\ref{tab:hm}. If paraphrases rated \textbf{1} or \textbf{2} are considered meaningful, we notice that SynPG generates meaningful paraphrases at a similar frequency to that of SIVAE. However, SynPG tends to generate more ungrammatical paraphrases (those rated $\textbf{1}$). We think the reason is that most of paraphrases generated by SIVAE are very similar to the source sentences, which are usually grammatically correct. On the other hand, SynPG is encouraged to use different syntactic structures from the source sentences to generate paraphrases, which may lead some grammatical errors.

Furthermore, we calculate the \emph{hit rate}, the percentage of generated paraphrases getting \textbf{2} and matching the target parse at the same time. The hit rate measures how often the generated paraphrases follow the target parses and preserve the semantics (verified by human evaluation) simultaneously. The results show that SynPG gets higher hit rate than other models.

\subsection{Target Parses vs. Target Templates}
\label{sec:tp}

Next, we discuss the influence of generating paraphrase by using templates instead of using full parse trees. For each paraphrase pair $(\mbf x_1, \mbf x_2)$ in test data, we consider two ways to generate the paraphrase. 1) Generating the paraphrase with the target parse. We use SynPG to generate a paraphrase directly from $(\mbf x_1, \mbf p_2)$. 2) Generating the paraphrase with the target template. We first use the parse generator to generate a parse $\tilde{\mbf p}$ from $(\mbf {tag}_1, \mbf t_2)$, where $\mbf {tag}_1$ is the tag sequence of $\mbf x_1$ and $\mbf t_2$ is the template of $\mbf p_2$. Then we use SynPG to generate a paraphrase from $(\mbf x_1, \tilde{\mbf p})$. We calculate the template matching accuracy to compare these two ways to generate paraphrases, as shown in Table~\ref{tab:tp_sc}. We also report the template matching accuracy of the generated parse $\tilde{\mbf p}$.

\begin{table}[t]
    \centering
    \small
    \aboverulesep = 0.5mm
    \belowrulesep = 0.5mm
    \begin{tabular}{lccccc}
        \toprule
        Model & \textbf{2} & \textbf{1} & \textbf{0} & \textbf{2+1} & Hit Rate\\
        \midrule
        BackTrans & 63.6 & 22.4 & 14.0 & 86.0 & 11.0 \\
        SIVAE     & 57.6 & 20.3 & 22.0 & 78.0 & 6.5 \\
        SynPG     & 44.3 & 32.0 & 23.7 & 76.3 & 28.9 \\
        \bottomrule
    \end{tabular}
    \caption{Human evaluation on a three-point scale (\textbf{0} = not a paraphrase, \textbf{1} = ungrammatical paraphrase, \textbf{2} = grammatical  paraphrase). SynPG performs better on hit rate (defined as the percentage of generated paraphrase getting $\textbf{2}$ and matching the target parse at the same time) than other unsupervised models. }
    \vspace{-1.2em}
    \label{tab:hm}
\end{table}

We find that most of generated parses $\tilde{\mbf p}$ indeed follow the target templates, which means that the parse generator usually generates good parses $\tilde{\mbf p}$. Next, we observe that generating paraphrases with target parses usually performs better than with target templates. The results show a trade-off. Using templates proves more effortless during the generation process, but may compromise the syntactic control ability. In comparison, by using the target parses, we have to provide more detailed parses, but our model can control the syntax better.

Another benefit of generating paraphrase with target templates is that we can easily generate a lot of syntactically different paraphrases by feeding the model with different templates. Table~\ref{tab:tptp} lists some paraphrases generated by SynPG with different templates. We can perceive that most generated paraphrases are grammatically correct and have similar meanings to the original sentence.

\begin{table}[t]
    \centering
    \small
    \aboverulesep = 0.5mm
    \belowrulesep = 0.5mm
    \begin{tabular}{ccccc}
        \toprule
        \multirow{2}{*}{Model} & \multicolumn{4}{c}{Template Matching Accuracy} \\
        \cmidrule{2-5}
        & ParaNMT & Quora & PAN & MRPC \\
        \midrule
        \scriptsize\makecell{Paraphrases \\ generated by \\ target parses} & 71.0 & 82.6 & 66.3 & 74.0 \\
        \midrule
        \scriptsize\makecell{Paraphrases \\ generated by \\ target templates} & 54.1 & 73.4 & 51.7 & 62.3 \\
        \midrule
        \scriptsize\makecell{Parses $\tilde{\mbf p}$ \\ generated by \\ parse generator} & 98.4 & 99.0 & 95.7 & 93.9 \\
        \bottomrule
    \end{tabular}
    \caption{Influence of using templates. Using templates proves more effortless during the generation process, but may compromise the syntactic control ability.}
    \label{tab:tp_sc}
    \vspace{-1.2em}
\end{table}

\subsection{Training SynPG on Larger Dataset}
Finally, we demonstrate that the performance of SynPG can be further improved and be even competitive to supervised models on some datasets if we consider more training data. The advantage of unsupervised paraphrase models is that we do not require parallel pairs for training. Therefore, we can easily boost the performance of SynPG by consider more unannotated texts into training.

We consider SynPG-Large, the SynPG model trained on the reference sentences of ParaNMT-50M as well as One Billion Word Benchmark \cite{Chelba14bword}, a large corpus for training language models. We sample about 24 million sentences from One Billion Word and add them to the training set. In addition, we fine-tune SynPG-Large on only the reference sentences of the testing paraphrase pairs, called SynPG-FT.

\begin{table*}[t]
    \centering
    \small
    \aboverulesep = 0.5mm
    \belowrulesep = 0.5mm
    \begin{tabular}{m{.4\textwidth}m{.5\textwidth}}
        \toprule
        Template & Generated Paraphrase \\
        \midrule
        Original & can you adjust the cameras? \\
        \texttt{(S(NP)(VP)(.))} & you can adjust the cameras. \\
        \texttt{(SBARQ(ADVP)(,)(S)(,)(SQ)(.))} & well, adjust the cameras , can you? \\
        \texttt{(S(PP)(,)(NP)(VP)(.))} &  on the cameras, you can adjust them? \\
        \midrule
        Original & she doesn't keep pictures from her childhood. \\
        \texttt{(SBARQ(WHADVP)(SQ)(.))} & 
why doesn't she keep her pictures from childhood. \\
        \texttt{(S(``)(NP)(VP)('')(NP)(VP)(.))} & 
`` she doesn't keep pictures from her childhood '' she said. \\
        \texttt{(S(ADVP)(NP)(VP)(.))} & 
perhaps she doesn't keep pictures from her childhood. \\
        \bottomrule
    \end{tabular}
    \caption{Paraphrases generated by SynPG with different templates.}
    \label{tab:tptp}
\end{table*}

\begin{table*}[t]
    \centering
    \small
    \aboverulesep = 0.5mm
    \belowrulesep = 0.5mm
    \begin{tabular}{x{.14\textwidth}y{.11\textwidth}x{.06\textwidth}x{.06\textwidth}x{.06\textwidth}x{.06\textwidth}x{.06\textwidth}x{.06\textwidth}x{.06\textwidth}x{.06\textwidth}}
        \toprule
        & \multirow{2}{*}{Model} & \multicolumn{2}{c}{ParaNMT} & \multicolumn{2}{c}{Quora} & \multicolumn{2}{c}{PAN} & \multicolumn{2}{c}{MRPC} \\
        \cmidrule{3-10}
        & & TMA & BLEU & TMA & BLEU & TMA & BLEU & TMA & BLEU \\
        \midrule
        \multirow{3}{*}{Ours} & SynPG & 71.0 & 32.2 & 82.6 & 33.2 & 66.3 & 26.4 & 74.0 & 26.2 \\
        & SynPG-Large & 70.3 & 31.8 & 83.8 & 34.7 & 66.6 & 27.1 & 79.3 & 36.2 \\
        & SynPG-FT & -- & -- & 86.3 & \textbf{44.4} & 66.4 & 34.2 & \textbf{80.7} & \textbf{44.6} \\
        \midrule
        \multirow{2}{*}{\makecell{Supervised\\ Models}}  & Seq2seq-Sup & 40.2 & 19.6 & 54.0 & 11.3 & 29.2 & 13.1 & 44.3 & 16.3 \\
        & SCPN & \textbf{83.9} & \textbf{58.3} & \textbf{87.1} & 41.0 & \textbf{72.3} & \textbf{37.6} & 80.1 & 41.8 \\
        \bottomrule
    \end{tabular}
    \caption{Training on larger dataset improves the performance of SynPG. Since training SynPG does not require annotated paraphrase pairs, it is possible to fine-tune SynPG on the texts in the target domain. With the fine-tuning, SynPG can have competitive or even better performance than supervised approaches.}
    \label{tab:ps_sup}
    \vspace{-1em}
\end{table*}

From Table~\ref{tab:ps_sup}, We observe that enlarging the training data set indeed improves the performance. Also, with the fine-tuning, the performance of SynPG can be much improved and even is better than the performance of supervised models on some datasets. The results demonstrate the potential of unsupervised paraphrase generation with syntactic control.

\subsection{Word Dropout Rate}
\label{sec:word}

\begin{figure}[!t]
	\centering
	\begin{subfigure}[b]{0.48\textwidth}
         \centering
         \includegraphics[width=.85\textwidth]{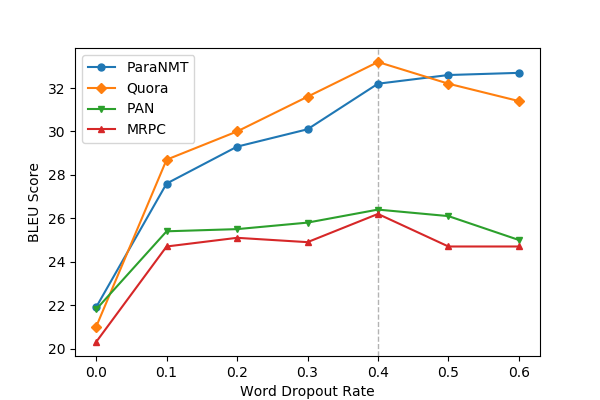}
         \caption{BLEU score}
         \label{fig:bleu}
     \end{subfigure}
     \begin{subfigure}[b]{0.48\textwidth}
         \centering
         \includegraphics[width=.85\textwidth]{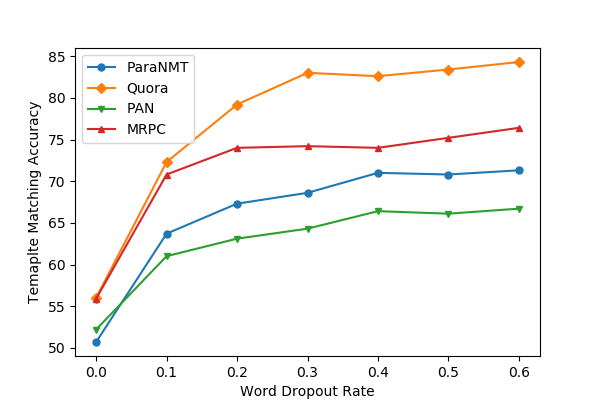}
         \caption{Template matching accuracy}
         \label{fig:tma}
     \end{subfigure}
     \caption{Influence of word drop out rate. Setting the word dropout rate to 0.4 can achieve the best BLEU score. However, higher word dropout rate leads to better template matching accuracy.}
     \label{fig:word}
     \vspace{-1.5em}
\end{figure}

The word dropout rate plays an important role for SynPG since it controls the ability of SynPG to generate new words in paraphrases. We test different word dropout rates and report the BLEU scores and the template matching accuracy in Figure~\ref{fig:word}.

From Figure~\ref{fig:bleu}, we can observe that setting the word dropout rate to 0.4 can achieve the best BLEU score in most of datasets. The only exception is ParaNMT, which is the dataset used for training. On the other hand, Figure~\ref{fig:tma} shows that higher word dropout rate leads to better template matching accuracy. The reason is that higher word dropout rate gives SynPG more flexibility to generate paraphrases. Therefore, the generated paraphrases can match the target syntactic specifications better. However, higher word dropout rate also make SynPG have less ability to preserve the meaning of source sentences. Considering all the factors above, we recommend to set the word dropout rate to 0.4 for SynPG.

\section{Improving Robustness of Models}
Recently, a lot of work show that NLP models can be fooled by different types of adversarial attacks \cite{Alzantot18attack,Ebrahimi18attack,Iyyer18scpn,Tan20attack, Jin20attack}. Those attacks generate \emph{adversarial examples} by slightly modifying the original sentences without changing the meanings, while the NLP models change the predictions on those examples. However, a robust model is expected to output the same labels. Therefore, how to make NLP models not affected by the adversarial examples becomes an important task.

Since SynPG is able to generate syntactically different paraphrases, we can improve the robustness of NLP models by data augmentation. The models trained with data augmentation are thus more robust to the syntactically adversarial examples \cite{Iyyer18scpn}, which are the adversarial sentences that are paraphrases to the original sentences but with syntactic difference. 

We conduct experiments on three classification tasks covered by GLUE benchmark \cite{Wang19glue}: SST-2, MRPC, and RTE. For each training example, we use SynPG to generate four syntactically different paraphrases and add them to the training set. We consider the setting to generate syntactically adversarial examples by SCPN \cite{Iyyer18scpn}. For each testing example, we generate five candidates of adversarial examples. If the classifier gives at least one wrong prediction on the candidates, we treat the attack to be successful.

We compare the model without data augmentation (Base) and with data augmentation (SynPG) in Table~\ref{tab:rb}. We observe that with the data augmentation, the accuracy before attacking is slightly worse than Base. However, after attacking, the percentage of examples changing predictions is much less than Base, which implies that data augmentation indeed improves the robustness of models.

\begin{table}[t]
    \centering
    \small
    \aboverulesep = 0.5mm
    \belowrulesep = 0.5mm
    \begin{tabular}{lcccccc}
        \toprule
        \multirow{2}{*}{Model} & \multicolumn{2}{c}{SST-2} & \multicolumn{2}{c}{MRPC} &  \multicolumn{2}{c}{RTE} \\
        \cmidrule{2-7}
        & Acc. & Brok. & Acc. & Brok. & Acc. & Brok. \\
        \midrule
        Base  & 91.9 & 46.7 & 84.1 & 52.8 & 63.2 & 58.3 \\
        SynPG & 88.9 & \textbf{39.6} & 80.1 & \textbf{35.5} & 60.7 & \textbf{33.9} \\
        \bottomrule
    \end{tabular}
    \caption{Data augmentation improves the robustness of models. SynPG denotes the base classifier trained on augmented data generated by SynPG. {Acc} denotes the accuracy in the original dataset (the higher is the better). Brok denotes the percentage of examples changing predictions after attacking (the lower is the better).}
    \label{tab:rb}
    \vspace{-1.2em}
\end{table}

\section{Related Work}
\paragraph{Paraphrase generation.}
Traditional approaches usually require hand-crafted rules, such as rule-based methods \cite{McKeown83rules}, thesaurus-based methods \cite{Bolshakov04thres,Kaucha06thres2}, and lattice matching methods \cite{Barzilay03lattice}.
However, the diversity of their generated paraphrases is usually limited.

Recently, neural models make success on paraphrase generation  \cite{Prakash16seq2seq1,Lapata17seq2seq2,Cao17seq2seq3, Egonmwan19seq2seq4,Li19seq2seq5,Gupta18seq2seq6}. These approaches treat paraphrase generation as a translation task and design seq2seq models based on a large amount of parallel data. To reduce the effort to collect parallel data, unsupervised paraphrase generation has attracted attention in recent years. \citet{Wieting17back,Gimpel18paranmt} use translation models to generate paraphrases via back-translation. \citet{Zhang19sivae}; \citet{Roy19vqvae} generate paraphrases based on variational autoencoders. Reinforcement learning techniques are also considered for paraphrase generation \cite{Li18rl}.

\paragraph{Controlled generation.}
Recent work on controlled generation can be grouped into two families. The first family is doing end-to-end training with an additional trigger to control the attributes, such as sentiment \cite{Shen17ctend1,Hu17end4,fu2018style,peng2018towards,DaiL19end3}, tense~\cite{Logeswaran18end2}, plots~\cite{Ammanabrolu2020StoryRE,Fan2019StrategiesFS,Tambwekar2019ControllableNS,yao2018plan,GoldfarbTarrant2019PlanWA,goldfarb2020content}, societal bias~\cite{wallace2019universal,sheng2020towards,sheng2020nice}, and syntax \cite{Iyyer18scpn,Goyal20preorder,Kumar20sgcp}.
The second family controls the attributes by learning disentangled representations. For example, \citet{Romanov19dis0} disentangle the meaning and the form of a sentence. \citet{Chen19dis1,Chen19dis2,Bao19dis3} disentangle the semantics and the syntax of a sentence.

\section{Conclusion}
We present syntactically controlled paraphrase generator (SynPG), an paraphrase model that can control the syntax of generated paraphrases based on the given syntactic specifications. SynPG is designed to disentangle the semantics and the syntax of sentences. The disentanglement enables SynPG to be trained without the need for annotated paraphrase pairs. Extensive experiments show that SynPG performs better syntactic control than unsupervised baselines, while the quality of the generated paraphrases is competitive to supervised approaches. Finally, we demonstrate that SynPG can improve the robustness of NLP models by generating additional training examples. SynPG is especially helpful for the domain where annotated paraphrases are hard to obtain but a large amount of unannotated text is available. One limitation of SynPG is the need for mannually providing target syntactic templates at inference time. We leave the automatic template generation as our future work.

\section*{Acknowledgments}
We thank anonymous reviewers for their helpful feedback. We thank Kashif Shah and UCLA-NLP group  for the valuable discussions and comments. This work is supported in part by Amazon Research Award.  

\clearpage

\bibliography{eacl2021}
\bibliographystyle{acl_natbib}

\end{document}